\PassOptionsToPackage{table}{xcolor}
\PassOptionsToPackage{normalem}{ulem}
\AddToHook{package/parskip/before}{\RequirePackage[subfigure]{tocloft}}
\documentclass[]{bytedance_seed}
\DeclareFontShape{T1}{bytesans}{b}{n}{<-> s * [1] seed/bytesans}{}
\DeclareFontShape{T1}{bytesans}{b}{it}{<-> s * [1] seed/bytesans}{}
\DeclareFontShape{T1}{bytesans}{b}{sc}{<-> s * [1] seed/bytesans}{}

\newlength{\seedheaderheight}
\AtBeginDocument{%
  \begingroup
    \pagestyle{firststyle}%
    \fancyhdrsettoheight{\seedheaderheight}{evenhead}%
    \global\seedheaderheight=\seedheaderheight
  \endgroup
  \addtolength{\headsep}{\headheight}%
  \setlength{\headheight}{\seedheaderheight}%
  \addtolength{\headsep}{-\headheight}%
}

\usepackage[toc,page,header]{appendix}

\usepackage{hyperref}
\usepackage{url}
\usepackage{xspace}
\usepackage{graphicx}
\usepackage{booktabs}
\usepackage{multirow}
\usepackage{xcolor}
\usepackage{amsmath,amssymb}
\usepackage{enumitem}
\usepackage{makecell}
\usepackage{pifont}
\newcommand{\nScenes}{590}
\newcommand{\nRooms}{3,037}
\newcommand{\nQuestions}{3,244}

\newcommand{\lenVid}{353.4 hrs}

\newcommand{\ourbench}{\textsc{ReSpaceBench}\xspace}
\newcommand{\ourbenchmini}{\textsc{ReSpaceBench}-\textit{mini}\xspace}
\newcommand{\best}{\cellcolor[HTML]{B7D2C4}}
\newcommand{\second}{\cellcolor[HTML]{DAE7DE}}

\title{Lost in the Hallway: Benchmarking Persistent Spatial Memory Across Revisits}

\author[1,2,*]{Henry Zheng}
\author[2]{Weibo Gong}
\author[1]{Yang Yue}
\author[1]{Rui Huang}
\author[2]{Yang Zhang}
\author[2, \dagger]{Gao Huang}

\affiliation[1]{Tsinghua University}
\affiliation[2]{PICO, ByteDance}

\contribution[*]{Work done at ByteDance}
\contribution[\dagger]{Corresponding author}

\abstract{
Persistent spatial understanding is essential for embodied agents to
operate reliably in environments they repeatedly encounter. It requires
recognizing revisited places despite viewpoint changes, distinguishing
them from similar-looking rooms, and updating memories as environments
change. Yet existing evaluations often leave physical place identity
across separated visits implicit. We introduce
\textsc{ReSpaceBench}, a benchmark comprising 3,244 questions across
590 synthetic and real-home scenes, organized into 9 tasks spanning
room association, persistent state tracking, and downstream reasoning.
Controlled walkthroughs define visits by physical room entry and
deliberately include same-type room distractors and cross-session object
changes. Our evaluation of 32 models and 3 agentic video-understanding
frameworks reveals substantial limitations: the strongest evaluated
zero-shot model achieves 50.5\% overall accuracy and 52.2\% on a
450-question subset, compared with 90.1\% for human annotators. The evaluated agentic frameworks offer no overall improvement over direct
inference with the same base model. Analysis of agent execution traces reveals evidence
omissions and incorrect cross-visit room associations as major failure
modes, suggesting that iterative evidence seeking alone does not ensure
consistent spatial memory. Further analyses expose a gap between
recognizing familiar rooms and identifying the correct visited room,
while annotation-guided sampling highlights the importance of preserving
intervening visits and transitions.
We also provide RS-30K, a separate training dataset from synthetic sources with no
template, scene, route, or video overlap with the benchmark.
Fine-tuning Qwen3-VL-8B raises overall accuracy from 35.5\% to 60.4\%,
with the largest gains in association and improvements extending to real-life scenes. \textsc{ReSpaceBench} and RS-30K
provide resources for diagnosing and improving persistent spatial memory
across revisits.
}

\firstauthor{Henry Zheng at \email{jh-zheng22@mails.tsinghua.edu.cn}}
\correspondence{Gao Huang at \email{gaohuang@tsinghua.edu.cn}}

\begin{document}
\maketitle

\section{Introduction}
\label{sec:intro}
\begin{figure}[ht]
    \centering
    \includegraphics[width=\linewidth]{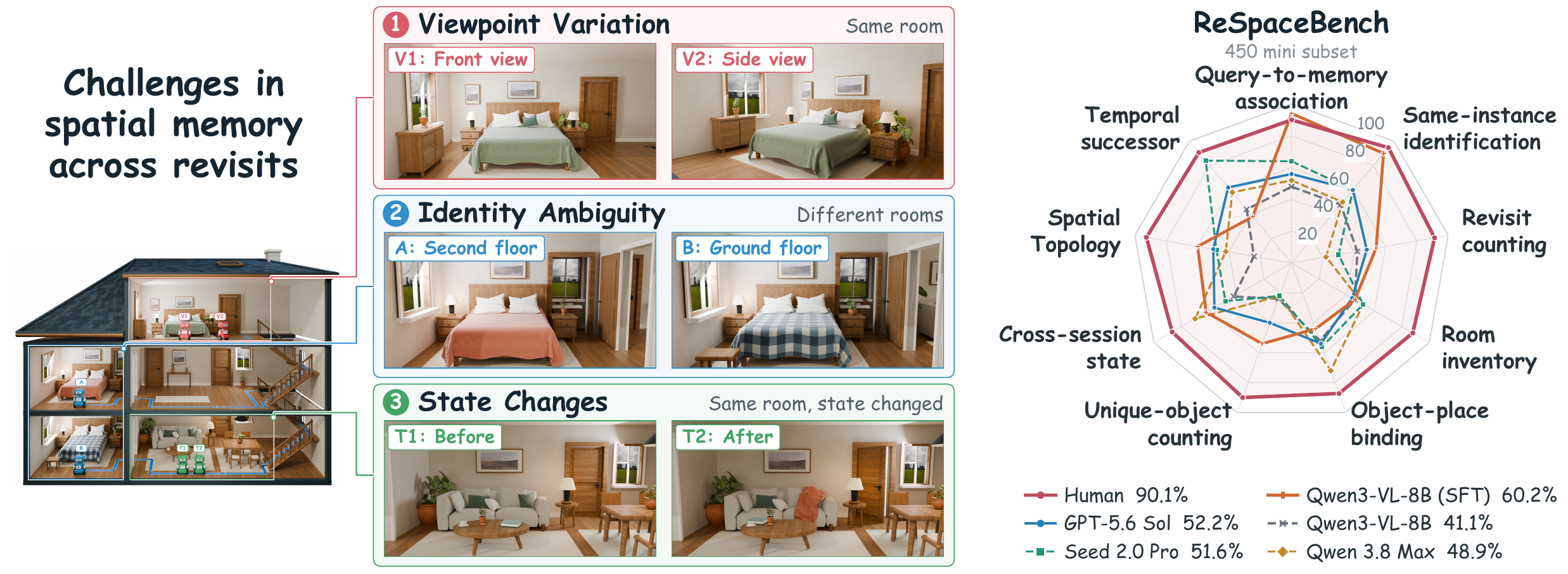}
    \caption{
    \textbf{Challenges in spatial memory across revisits.}
    \textbf{Left:} Three challenges arise when revisiting a previously explored environment:
    (1) \emph{Viewpoint Variation}: recognizing the same room from different viewpoints;
    (2) \emph{Identity Ambiguity}: distinguishing between visually similar but distinct rooms; and
    (3) \emph{State Changes}: tracking changes within the same room across repeated visits.
    \textbf{Right:} Human and MLLM accuracy across all 9 tasks of \ourbenchmini{} (450 questions).
    Even the best-performing model remains substantially below human performance.
    }
    \label{fig:teaser}
\end{figure}
Embodied agents repeatedly encounter the same physical environments during everyday activities~\citep{hawes2016strands,kunze2018artificial}.
A robot or wearable assistant may return to a room from a different viewpoint, after visiting other spaces, or after its contents have changed.
To build on past experience during these revisits, it must associate new observations with the correct physical place and update its spatial memory accordingly~\citep{schmid2024Khronos,labbe2019rtabmap,hughes2022hydra}.
Yet \emph{viewpoint variation} can make even a familiar room appear substantially different across visits, complicating place recognition~\citep{garg2021your,lowry2016vpr,zaffar2021vpr}.
Conversely, different rooms may look alike, creating \emph{identity ambiguity} about the correct identity of the room being revisited~\citep{lowry2016vpr,zhang2021visual}.
Over time, the agent must also track \emph{state changes} in the room's contents and update the corresponding memory~\citep{kurenkov2023modeling,narayana2020lifelong}.
Revisits therefore test whether systems can preserve place identity while tracking changes across temporal observations (Fig.~\ref{fig:teaser}).

Recent benchmarks have expanded the evaluation of multimodal large language
models (MLLMs) from single-image 3D relations to video-based spatial reasoning
and long-horizon memory
\citep{ma20253dsrbench,yang2025thinking,lin2025mmsivideobench,lang2026longspace}.
Yet models still struggle with global cognitive maps, cross-view
correspondence, and coherent spatial memory
\citep{yang2025thinking,yeh2026seeing,liao2026spamem,lin2025mmsivideobench}.
Cross-visit place association, a task well studied in robotics through loop closure
and visual place recognition \citep{cadena2016past,lowry2016vpr}, remains
largely implicit in the spatial understanding task evaluations. Recent benchmarks, UCS-Bench~\citep{wang2026ucsbench} and SpaMEM~\citep{liao2026spamem}
evaluate persistent spatial reasoning and memory under scene changes,
while LongSpace-Bench~\citep{lang2026longspace} covers scene consistency
and long-horizon recall; however, none establishes a dedicated protocol for room
re-identification across controlled revisits.
OVO-S-Bench~\citep{li2026ovosbench} includes revisit detection and counting,
but defines visits by whether a location enters the field of view, allowing
a room briefly glimpsed during a camera turn to count without physical entry.
Its construction also does not systematically control same-type rooms as
competing identities. Thus, associating observations with specific physical
rooms across controlled revisits remains under-explored.

We present \ourbench{}, a benchmark for persistent spatial understanding
across revisits. Its 9 tasks span three task families:
\emph{association}, recognizing previously visited rooms and matching them
to earlier observations; \emph{persistent state}, maintaining and updating
room and object memories; and \emph{downstream reasoning}, recovering
temporal order and spatial topology. Walkthroughs follow controlled routes,
with visits defined by physical room entry rather than visual observation.
We select homes with spaces of the same type to serve
as identity distractors, testing whether models recognize a specific place
rather than merely remembering its semantic category.

We construct \ourbench{} from synthetic homes in 3D-FRONT~\citep{fu20213dfront} and ProcTHOR~\citep{deitke2022procthor}, real homes scanned with FARO
laser scanners~\citep{faro_focus}, and egocentric videos recorded with
a PICO headset~\citep{pico}. Using annotated scene metadata and constructed navigation meshes, we
use our designed question templates to generate task-conditioned routes: each template specifies both the
question and the navigation requirements needed to answer it. The resulting
route trajectories include egocentric
videos with repeated room visits. We generate answers from metadata, then apply rule-based checks with human verification to remove
question--answer pairs that are inconsistent with the video. Combining synthetic environments and real
homes, \ourbench comprises \nScenes{} scenes, \nRooms{} rooms, \lenVid,
and \nQuestions{} questions.

Our evaluation of 32 models and three agentic video-understanding frameworks
(Section~\ref{sec:results}) reveals substantial limitations in persistent
spatial memory. Proprietary models remains the strongest zeroshot models in our evaluation, while
specialized spatial and video models do not consistently outperform
open-source general-purpose MLLMs. The evaluated agentic frameworks offer
no overall gains over direct inference. On a shared 450-question subset,
humans achieve 90.1\% accuracy
compared with 52.2\% for the strongest evaluated zero-shot model GPT-5.6 Sol. Further analysis of the six proprietary models on
query-to-memory association shows that wrong-room selections account
for 66.5--88.3\% of their errors, exposing a gap between recognizing a room
as familiar and identifying the correct visited room. To support model
development, we also provide 30k training questions from ProcTHOR training
scenes with no scene, route, or video overlap with the benchmark.
Fine-tuning Qwen3-VL-8B on this dataset raises overall accuracy from
35.5\% to 60.4\%, with the largest gains in association and smaller,
less consistent gains in state tracking and reasoning.

Our contributions are as follows.
\begin{itemize}[leftmargin=*,itemsep=1pt,topsep=2pt]
  \item \textbf{A benchmark for persistent spatial memory.}
  We introduce \ourbench{}, comprising 9 tasks in 3 task families across synthetic
  and real homes. Controlled revisits, same-type room distractors, and
  cross-session state changes support explicit evaluation of room
  association, persistent state tracking, and downstream spatial reasoning.

  \item \textbf{A systematic evaluation and error analysis.}
  We evaluate 32 models and 3 agentic video-understanding frameworks,
  with a human performance reference on a shared subset. Detailed analyses of room-association
  errors, agentic failures, and evidence coverage identify recurring
  limitations in maintaining and using spatial memory across revisits.

  \item \textbf{A training dataset.}
  We release 30k training samples generated from ProcTHOR scenes,
  with no template, scene, route or video overlap with the benchmark.
  A targeted fine-tuning study demonstrates their utility for developing
  models with stronger persistent spatial memory.
\end{itemize}

\newcommand{\benchyes}{\ding{51}}
\newcommand{\benchno}{\ding{55}}
\newcommand{\benchpartial}{\ensuremath{\triangle}}

\begin{table}[t]
\centering
\caption{
Comparison with related spatial understanding and memory benchmarks.
Symbols reflect explicit evaluation targets:
\benchyes{} indicates explicit coverage;
\benchno{} indicates a criterion not explicitly targeted;
\benchpartial{} indicates related coverage that does not clearly establish
the criterion.
\emph{Controlled revisits} refers to planned room re-entries along
multi-room trajectories.
\emph{Same-type room distractors} refers to distinct rooms of the same
category deliberately included as competing place identities.
}
\label{tab:benchmark_comparison}
\vspace{5pt}
\small
\setlength{\tabcolsep}{3pt}
\renewcommand{\arraystretch}{1.15}

\resizebox{\textwidth}{!}{%
\begin{tabular}{@{}llccccc@{}}
\toprule
\textbf{Benchmark}
& \textbf{Input}
& \multicolumn{3}{c}{\textbf{Evaluation coverage}}
& \multicolumn{2}{c}{\textbf{Revisit-focused design}} \\
\cmidrule(lr){3-5}
\cmidrule(lr){6-7}
&
& \makecell{Spatial\\memory}
& \makecell{State\\changes}
& \makecell{Revisit\\recognition}
& \makecell{Controlled\\revisits}
& \makecell{Same-type room\\distractors} \\
\midrule

VSI-Bench \citep{yang2025thinking}
& Video
& \benchyes & \benchno & \benchno
& \benchno & \benchno \\

VSI-SUPER (Cambrian-S) \citep{yang2026cambrian}
& Video
& \benchyes & \benchno & \benchno
& \benchno & \benchno \\

All-Angles Bench \citep{yeh2026seeing}
& Multi-view Img.
& \benchpartial & \benchno & \benchno
& \benchno & \benchno \\

SpaCE-10 \citep{space10}
& Multi-view Img.
& \benchpartial & \benchno & \benchno
& \benchno & \benchno \\

MMSI-Video-Bench \citep{lin2025mmsivideobench}
& \makecell[l]{Video}
& \benchyes & \benchyes & \benchno
& \benchno & \benchno \\

SpaMEM \citep{liao2026spamem}
& Img. Sequence
& \benchyes & \benchyes & \benchno
& \benchno & \benchno \\

UCS-Bench \citep{wang2026ucsbench}
& Video
& \benchyes & \benchyes & \benchno
& \benchno & \benchno \\

SAW-Bench \citep{sawbench}
& Video
& \benchyes & \benchyes & \benchno
& \benchno & \benchno \\

LongSpace-Bench \citep{lang2026longspace}
& Video
& \benchyes & \benchyes & \benchpartial
& \benchno & \benchno \\

OVO-S-Bench \citep{li2026ovosbench}
& Video
& \benchyes & \benchyes & \benchyes
& \benchno & \benchno \\

\midrule
\textbf{\ourbench{} (ours)}
& Video
& \benchyes & \benchyes & \benchyes
& \benchyes & \benchyes \\
\bottomrule
\end{tabular}%
}

\end{table}

\section{Related Work}
\label{sec:related-work}

\subsection{Long-context visual retrieval.}
Long-context evaluations examine whether models can retrieve and combine sparse
visual evidence from extended multimodal inputs. Visual Haystacks tests visual needle
retrieval and cross-image reasoning, revealing sensitivity to evidence position
\citep{wu2024visualhaystacks}. MMLongBench broadens evaluation to diverse
interleaved vision--language tasks at controlled context lengths
\citep{wang2025mmlongbench}, while Video-LevelGauge systematically probes
positional bias in video contexts \citep{xia2025videolevelgauge}.
Recently, agentic video-understanding frameworks such as AVP, DVD, and VideoAgent
use tools to retrieve and inspect video evidence
\citep{wang2026avp, zhang2026dvd, wang2024video}.
Retrieving relevant observations, however, does not establish whether they refer to the same physical room across separate visits. \ourbench{} explicitly
evaluates this cross-visit room association over long visual contexts.

\subsection{Spatial intelligence and long-horizon memory.}
Recent benchmarks address complementary aspects of spatial intelligence and
long-horizon memory. SpaMEM studies spatial belief updates and object-identity
continuity under action-conditioned scene changes in ProcTHOR
\citep{liao2026spamem}, while UCS-Bench evaluates continual spatial reasoning
relative to users' changing locations in egocentric streams
\citep{wang2026ucsbench}. EgoMemReason probes entity, event, and behavioral
memory over week-long videos \citep{wang2026egomemreason}, and 3DMem-Bench
evaluates memory-dependent question answering and embodied tasks in dynamic
3D environments \citep{hu20253dllmmem}. MMSI-Video-Bench includes cross-video
spatial reasoning \citep{lin2025mmsivideobench}.
All-Angles Bench~\citep{yeh2026seeing} and SpaCE-10~\citep{space10}
evaluate multi-view and compositional spatial understanding of a static scene
from several images, and SAW-Bench~\citep{sawbench} evaluates situated awareness in egocentric videos, including changes between two videos
of the same scene. However, these benchmarks do not systematically evaluate whether models can recognize a physical room when it is revisited.
In generative world modeling,
MemoBench evaluates whether video generators recover objects in their
updated states after disappearance and reappearance
\citep{chen2026memobench}. More closely related to our setting,
LongSpace-Bench includes scene consistency, state change, and route recall
in long room-tour videos \citep{lang2026longspace}. OVO-S-Bench evaluates
scene revisit detection and counting under a causal, prefix-only protocol,
with visits in its counting task defined by a location entering the field
of view \citep{li2026ovosbench}. \ourbench{} defines visits by physical room
entry and deliberately includes distinct same-type rooms as competing
identities along controlled revisit trajectories.

\subsection{Persistent association and structured spatial memory.}
Persistent association underlies place recognition and loop closure in SLAM
\citep{cadena2016past}. REMIND uses appearance, spatial context, and joint
assignment for indoor object re-identification \citep{diazpereda2026remind},
while R4DSG indexes egocentric memory by time, place, persistent objects, and
anchor-relative changes \citep{ma2026r4dsg}. PBD-AG separates a robot-verified
structural baseline from revisable dynamic events \citep{bao2026pbd}, and
GESTO uses event context within a persistent 4D scene graph to refine object
associations \citep{bartoli2026gesto}. Complementing these methods,
\ourbench{} evaluates whether MLLMs can associate observations with
persistent room identities and use visit histories and state information
to answer questions across revisits.

\begin{figure}[t]
  \centering
  \includegraphics[width=\linewidth]{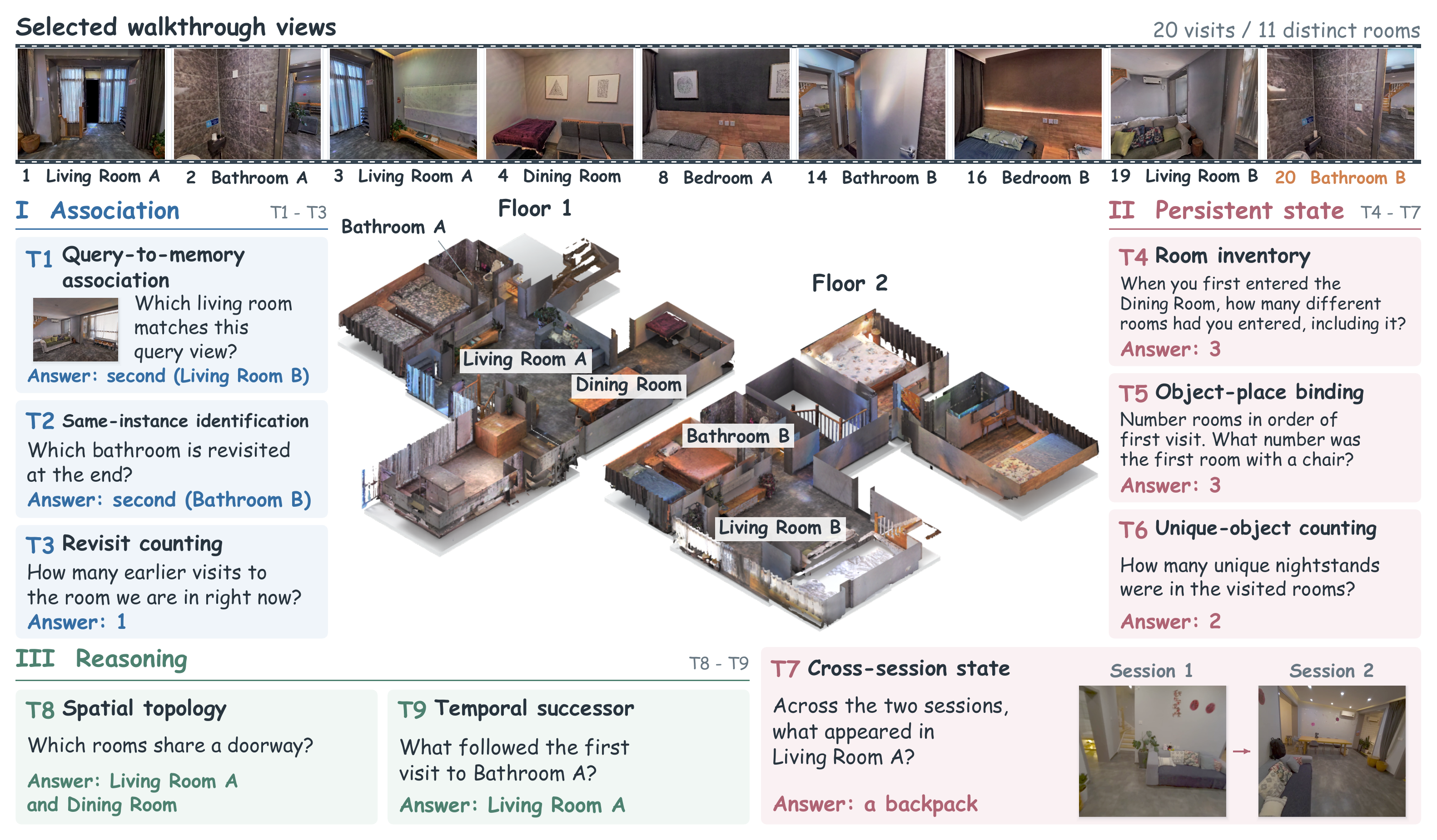}
  \vspace{-10pt}
  \caption{\textbf{Benchmark taxonomy:} \ourbench{} comprises 9 tasks across association (T1--T3), persistent state (T4--T7), and reasoning (T8--T9). Representative questions and answers illustrate
    these tasks through room revisits, same-type room distractors, and
    cross-session state changes.}
  \label{fig:tasks}
\end{figure}

\section{\ourbench}
\label{sec:bench}
\label{sec:tasks}

\ourbench evaluates persistent spatial memory through egocentric walkthrough that
return to previously visited places. Its tasks span three task families
(Fig.~\ref{fig:tasks}): \emph{association}, recognizing a revisited room and matching
it to an earlier observation; \emph{persistent state}, maintaining room inventories,
object--place bindings, and object states across visits; and \emph{reasoning}, inferring
temporal order and spatial connectivity from this history. The benchmark evaluates these capabilities using rendered walkthroughs of synthetic and FARO-scanned homes, together with PICO headset recordings of real homes.

\paragraph{\textbf{Task overview.}}
For \emph{association},
T1 (\emph{query-to-memory association}) matches a query room to a visited room or labels it as never entered;
T2 (\emph{same-instance identification}) identifies which previously
visited same semantic type room is being revisited;
and T3 (\emph{revisit counting}) counts prior entries into a specified room.
For \emph{persistent state},
T4 (\emph{room inventory}) counts distinct visited rooms, overall
or by semantic type, or determines a room's position in the order
of first visits;
T5 (\emph{object--place binding}) reports first-visit of the first room containing an object of a queried type;
T6 (\emph{unique-object counting}) counts distinct objects of a queried type across revisits;
and T7 (\emph{cross-session state}) tracks object locations and changes across sessions in the same home.
For \emph{reasoning},
T8 (\emph{spatial topology}) identifies room pairs directly connected by a doorway,
and T9 (\emph{temporal successor}) identifies the room type entered immediately after a specified visit.

\subsection{Benchmark Construction Pipeline}
\label{sec:pipeline}

\begin{figure}[t]
  \centering
  \includegraphics[width=\linewidth]{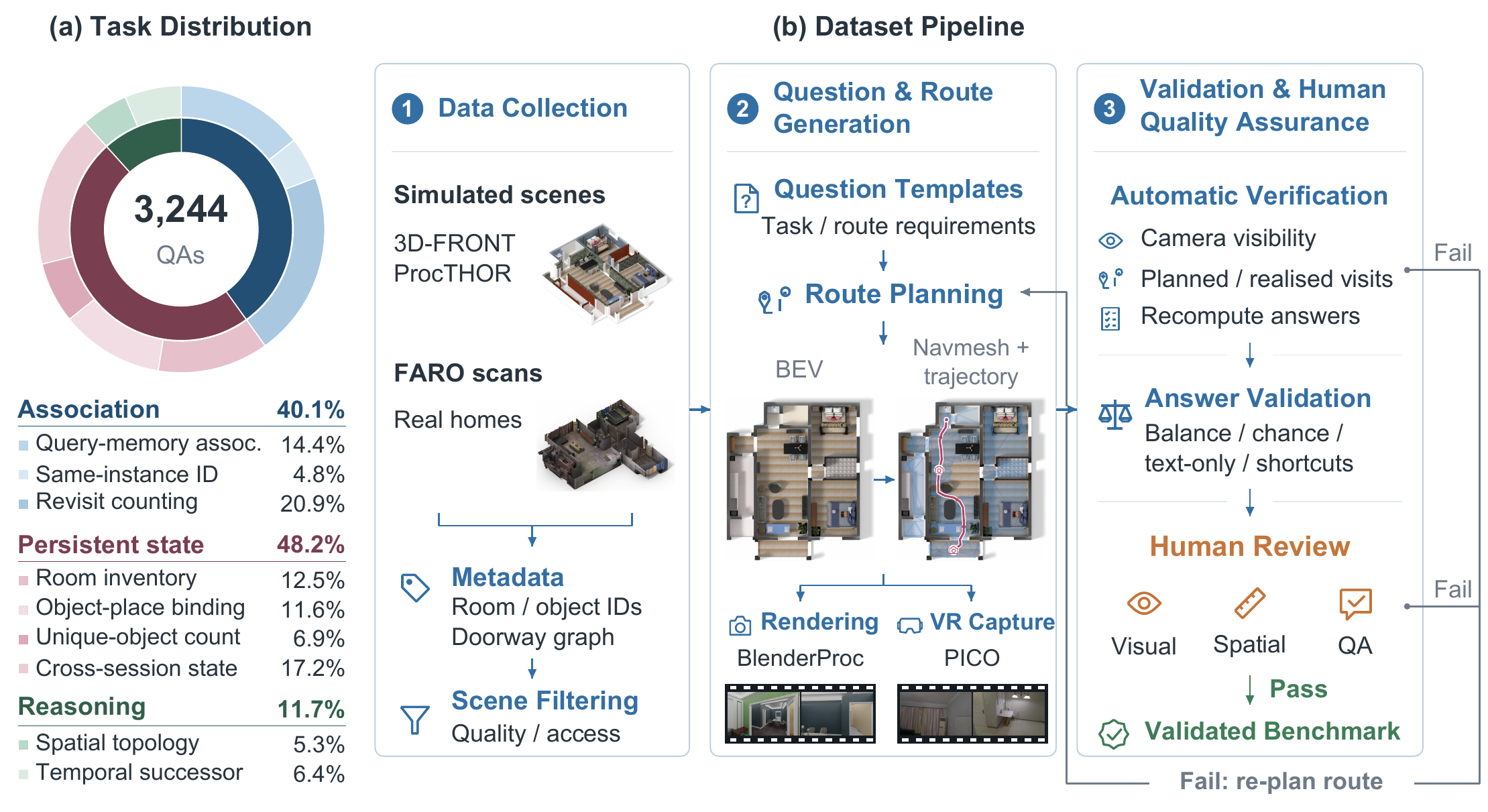}
  \caption{Benchmark statistics and construction pipeline. \textbf{(a)} Question distribution across the 3 task families and 9 tasks. \textbf{(b)} The 3-stage construction pipeline for synthetic and real homes, with automated checks and human quality control. The camera path is illustrative.}
  \label{fig:pipeline}
\end{figure}

Fig.~\ref{fig:pipeline}(b) shows the 3 stages of benchmark construction. Scene
metadata and camera trajectories connects each question to the rooms and objects observed in its video, providing
a shared and consistent basis for both question generation and answer verification.

\paragraph{\textbf{Scene collection, filtering, and metadata.}}
\label{sec:real}
To ensure that the benchmark poses meaningful spatial-memory challenges, we select large homes with diverse room layouts and multiple rooms of the same semantic type. We source synthetic household environments from 3D-FRONT \citep{fu20213dfront}, which provides professionally designed furnished apartments, and ProcTHOR \citep{deitke2022procthor}, which contains procedurally generated homes. To complement these synthetic scenes, we scan real-world residential homes using a FARO laser scanner and record egocentric activities with a PICO headset. For synthetic scenes, we extract room geometry and semantic types,
persistent room and object identities, object categories and
locations, and doorway connectivity from the source metadata. For real homes, we manually annotate the corresponding information
in the registered 3D scenes and align the annotations with the
recorded egocentric trajectories.

\paragraph{\textbf{Route planning and egocentric video generation.}}
We plan walkthrough routes using room layout to balance
room coverage and controlled revisits. Task-specific configurations include visits to visually similar rooms of the same semantic type, episodes spanning single and multiple homes,
and repeated sessions in the same home with controlled object changes.
For both synthetic and scanned homes, we construct navigation meshes using Habitat Simulator~\citep{savva2019habitat}. Then, we adapt the navmesh-based navigation setup of Pred-EQA~\citep{yuan2026predeqa} to generate traversable paths that follow the planned room-entry sequences. We convert each path into a continuous camera trajectory
and render egocentric videos using
BlenderProc~\citep{denninger2023blenderproc}, then concatenating videos across multiple homes when needed. Rendered walkthroughs use nominal walking speeds of 0.6--0.8\,m/s. At each room entry, the camera pauses for a $270^\circ$ look-around lasting 6--9 seconds, providing sustained observations of the room. Hallways and stairwells are treated as transit spaces and are not counted as room visits. Each video is paired with camera poses, timestamps, room-entry events, and object-state records to support question generation, answer verification.

\paragraph{\textbf{Question generation from templates.}}
We instantiate task-specific question templates using scene
metadata, ordered room-entry sequences, and object-state records. The templates cover room association, room and object inventories, object to place binding, state changes, temporal order, and room connectivity. Rooms are referred to by semantic type when the type is unique within a scene. Otherwise, we disambiguate them by order of first visit or unique visual cues (e.g., \emph{the second bedroom visited} or \emph{the bedroom with wooden floor}). Ground-truth answers are deterministically derived from the corresponding scene annotations and realized trajectory. For multiple-choice questions, distractors are included to reflect plausible
spatial-memory errors, such as confusing distinct rooms of the same
type or inferring direct doorway connectivity from consecutive visits.

\paragraph{\textbf{Automatic and Human Quality Control.}}
We verify each walkthrough by comparing the realized room-entry sequence with the planned route, rejecting trajectories with missing or unintended room entries. Ground-truth answers are then recomputed from the verified entry sequence and object-state records, and questions involving ambiguous visits or insufficient visual evidence are discarded. We further combine task-specific automated checks with manual inspection to assess rendering quality, room and object visibility, and consistency between the video and its metadata.
During construction, shortcut diagnostics is implemented to examine answer-frequency biases, question and option wording, and
room-type priors.
These checks guide revisions to question templates and answer balancing,
while blind and frequency baselines provide additional diagnostics
on the evaluation set.

\paragraph{\textbf{Training data.}}
To support model development, we additionally provide RS-30K,
a training dataset of 30k questions generated from ProcTHOR
training scenes. RS-30K uses a separate set of question templates and shares no
scenes, routes, or videos with \ourbench{}.
It provides targeted supervision without reusing evaluation
environments or walkthroughs. We assess its utility through supervised fine-tuning of
Qwen3-VL-8B, with results reported in Sec.~\ref{sec:results}.

\subsection{Benchmark Statistics}
\label{sec:statistics}

\ourbench{} contains 3,244 evaluation questions, including 3,026 questions derived from synthetic environments and 218 questions from real homes.
The real-home questions use 35 walkthrough videos rendered from FARO-scanned homes, which imitate egocentric videos by following the camera height, field of view, and motion of PICO recordings, and 13 videos from PICO headset recordings. Because a FARO scan captures a static scene, the PICO recordings are used for T7 (cross-session state), where objects change between sessions.
Fig.~\ref{fig:pipeline}(a) shows the distribution across
3 task families and 9 tasks.
We additionally use \ourbenchmini{}, a 450-question subset containing 50 samples per task, for human evaluation, agentic experiments, and frame-sampling analyses (Sec.~\ref{sec:results}). RS-30K is a separate training dataset and is therefore excluded from these evaluation statistics.

\subsection{Evaluation Protocol}
\label{sec:protocol}

Each model receives the walkthrough video and a textual question,
together with textual answer options for multiple-choice items.
T1 additionally provides a short query clip showing a room from
a new viewpoint. The model must match this room to a previously
entered room or select the ``not entered'' option. Scene metadata, camera poses, and ground-truth event annotations are only for further analysis and not provided to models under the default evaluation protocol.

Responses use one of two formats: an option letter for
\emph{multiple-choice} questions or a value for \emph{numeric}
questions.
Multiple-choice responses are scored by whether the selected
option is correct, while numeric responses are scored by exact
match to the ground-truth value. Overall accuracy is computed as the unweighted mean of the nine task accuracies.
We report accuracy for each task and the average of each task family, alongside
the Blind, Chance, and Frequency baselines described in
Section~\ref{sec:results}.

\section{Evaluations on \ourbench{}}
\label{sec:results}

We evaluate 32 models covering proprietary MLLMs, open-source
general-purpose MLLMs, unified models, and specialized spatial and
video-understanding models, as well as 3 agentic video-understanding
frameworks. Except for targeted fine-tuning, models use their standard
inference interfaces without benchmark-specific training.
For direct inference, we sample videos at 1\,fps by default, retaining
approximately 6--9 views per room-entry look-around, and further
downsample when required by model-specific context limits.
Details are provided in the appendix.

\begin{table}
\setlength{\fboxsep}{0pt}
\caption{Evaluation of 32 models on \ourbench{}. \protect\colorbox[HTML]{B8E4C2}{Darker green} entries denote the best among proprietary models and \protect\colorbox[HTML]{E0F1E3}{lighter green} entries denote the best among open-source models. Avg.\ under each task family is the mean of its task accuracies. $^\dagger$Qwen3-VL-8B fine-tuned on RS-30K training data generated from the ProcTHOR training set.}
\vspace{5pt}
\centering
\small
\setlength{\tabcolsep}{3.5pt}
\begin{tabular}{l  c c  cccc  ccccc  ccc}
\toprule
\multirow{2}{*}{Methods} & \multirow{2}{*}{Rank}&\multirow{2}{*}{Avg.}  & \multicolumn{4}{c}{Association} & \multicolumn{5}{c}{Persistent state} & \multicolumn{3}{c}{Reasoning} \\
\cmidrule(lr){4-7}
\cmidrule(lr){8-12}
\cmidrule(lr){13-15}
 &  && T1 & T2 & T3 & Avg. & T4 & T5 & T6 & T7 & Avg. & T8 & T9 & Avg. \\
\midrule
\multicolumn{14}{l}{\textit{Baseline}} & \\
\addlinespace[1pt]
Chance Level (Random) & -- &-- & 25.0 & 33.3 & -- & -- & -- & -- & -- & 25.0 & -- & 25.0 & 25.0 & -- \\
Chance Level (Frequency) & -- &-- & 25.1 & 34.0 & 25.9 & -- & 27.8 & 22.4 & 32.1 & 25.5 & -- & 26.7 & 25.7 & -- \\
Blind (Seed 2.0 Pro~\citep{bytedance2026seed20}) & -- & 27.4 & 23.4 & 37.0 & 25.1 & 28.5 & 25.1 & 19.2 & 25.4 & 21.4 & 22.8 & 33.1 & 36.8 & 35.0 \\
\addlinespace[2pt]
\midrule
\addlinespace[1pt]
\multicolumn{14}{l}{\ourbench{}-\textit{mini}} & \\
GPT-5.6 Sol~\citep{openai2026gpt56} & -- & 52.2 & 56.0 & 60.0 & 48.0 & 54.7 & 44.0 & 54.0 & 40.0 & 56.0 & 48.5 & 50.0 & 62.0 & 56.0 \\
Human (Mean of 3) & -- & 90.1 & 90.0 & 94.7 & 91.3 & 92.0 & 88.0 & 87.3 & 90.0 & 86.7 & 88.0 & 92.7 & 90.7 & 91.7 \\
\addlinespace[1pt]
\addlinespace[2pt]
\midrule
\addlinespace[1pt]
\multicolumn{14}{l}{\textit{Proprietary Models}} & \\
\addlinespace[1pt]
GPT-5.6 Sol~\citep{openai2026gpt56} & 1 & \best{}50.5 & 56.2 & \best{}65.4 & \best{}34.2 & \best{}51.9 & \best{}55.4 & \best{}58.4 & 30.0 & 47.0 & 47.7 & 41.9 & 66.2 & 54.0 \\
GPT-5.6 Luna~\citep{openai2026gpt56} & 17 & 32.5 & 38.6 & 44.2 & 14.0 & 32.3 & 31.8 & 39.0 & \best{}32.7 & 34.9 & 34.6 & 25.0 & 32.4 & 28.7 \\
Seed 2.1 Pro~\citep{bytedance2026seed21} & 3 & 48.0 & \best{}58.8 & 59.0 & 26.7 & 48.2 & 48.3 & 54.4 & 30.0 & 46.3 & 44.8 & \best{}44.2 & 64.7 & 54.5 \\
Seed 2.0 Pro~\citep{bytedance2026seed20} & 2 & 49.3 & 52.1 & 60.3 & 29.1 & 47.2 & 48.8 & 52.3 & 31.8 & 49.7 & 45.6 & 43.0 & \best{}76.3 & \best{}59.7 \\
Qwen 3.8 Max~\cite{qwen2026qwen38max} & 4 & 46.3 & 50.0 & 49.4 & 18.9 & 39.4 & 48.5 & 56.0 & 27.4 & \best{}61.7 & \best{}48.4 & 43.0 & 62.3 & 52.7 \\
Gemini 3.1 Pro~\cite{deepmind2026gemini31pro} & 12 & 34.0 & 44.2 & 39.7 & 19.5 & 34.5 & 25.4 & 36.9 & 19.7 & 44.0 & 31.5 & 30.8 & 45.4 & 38.1 \\
\addlinespace[2pt]
\midrule
\addlinespace[1pt]
\multicolumn{14}{l}{\textit{Open-source General Purpose Models}} & \\
\addlinespace[1pt]
Qwen3.6-27B~\citep{qwen3.6-27b} & 6 &40.3 & 34.5 & 40.4 & 27.4 & 34.1 &  \second{}42.6 &  \second{}55.4 & 33.2 & \second{} 48.8 & \second{}45.0 & 27.3 & 53.1 & 40.2 \\
Qwen3.6-35B-A3B~\citep{qwen36_35b_a3b} & 8 &36.8 & 32.0 & \second{}48.7 & 22.6 & 34.4 & 38.2 & 35.3 & 25.6 & 40.8 & 35.0 & 32.6 & 55.6 & 44.1 \\
Qwen3-VL-32B~\citep{bai2025qwen3vl} & 5 & \second{}41.3 & \second{}45.1 & 44.9 & \second{}30.1 & \second{}40.0 & 36.5 & 49.3 & 29.6 & 41.5 & 39.2 & \second{}36.6 & \second{}58.0 & \second{}47.3 \\
Qwen3-VL-8B~\citep{bai2025qwen3vl} & 10 &35.5 & 43.1 & 36.5 & 23.6 & 34.4 & 38.4 & 42.7 & 26.9 & 41.5 & 37.4 & 25.6 & 41.5 & 33.6 \\
Qwen3-VL-4B~\citep{bai2025qwen3vl} & 16 &32.7 & 39.1 & 35.9 & 24.5 & 33.1 & 26.1 & 32.1 & 30.9 & 32.2 & 30.3 & 28.5 & 45.4 & 36.9 \\
InternVL3.5-38B~\citep{wang2025internvl3} & 7 &38.0 & 44.2 & 40.4 & 25.5 & 36.7 & 31.3 & 31.8 & 34.1 & 44.5 & 35.4 & 34.9 & 55.6 & 45.2 \\
InternVL3.5-30B-A3B~\citep{wang2025internvl3} & 19 & 30.7 & 28.3 & 39.7 & 24.2 & 30.8 & 35.5 & 24.1 & 29.1 & 38.5 & 31.8 & 24.4 & 32.4 & 28.4 \\
InternVL3.5-8B~\citep{wang2025internvl3} & 21 &30.4 & 25.8 & 32.7 & 26.1 & 28.2 & 29.3 & 23.1 & 29.6 & 39.4 & 30.3 & 27.3 & 40.6 & 34.0 \\
GLM-4.6V-106B~\citep{zai2025glm46v} & 13 & 33.1 & 38.8 & 37.2 & 26.0 & 34.0 & 32.8 & 12.5 & 32.7 & 47.6 & 31.4 & 26.2 & 44.4 & 35.3 \\
GLM-4.6V-Flash-9B~\citep{zai2025glm46v} & 22 & 29.0 & 42.3 & 34.0 & 21.1 & 32.4 & 23.2 & 15.4 & 17.0 & 40.3 & 24.0 & 24.4 & 43.5 & 33.9 \\
Gemma-4 26B-A4B~\citep{gemmateam2026gemma4} & 15 & 32.8 & 29.8 & 35.3 & 25.7 & 30.2 & 32.3 & 34.5 & 26.5 & 39.0 & 33.1 & 28.5 & 44.0 & 36.2 \\
Gemma-4 E4B~\citep{gemmateam2026gemma4} & 29 & 25.0 & 33.5 & 35.9 & 26.5 & 32.0 & 13.3 & 27.3 & 0.4 & 31.1 & 18.0 & 20.9 & 35.7 & 28.3 \\
Gemma-4 E2B~\citep{gemmateam2026gemma4} & 30 & 24.2 & 26.4 & 30.8 & 26.4 & 27.9 & 20.4 & 22.5 & 16.1 & 16.3 & 18.9 & 23.3 & 35.7 & 29.5 \\
Cosmos3-Super-64B~\citep{agarwal2026cosmos} & 13 & 33.1 & 40.8 & 35.3 & 21.7 & 32.6 & 24.9 & 38.7 & 27.8 & 38.5 & 32.5 & 29.1 & 41.5 & 35.3 \\
\addlinespace[2pt]
\midrule
\addlinespace[1pt]
\multicolumn{14}{l}{\textit{Video Understanding Models}} & \\
\addlinespace[1pt]
Video-XL-2-8B~\citep{qin2025videoxl2} & 24 & 28.7 & 25.5 & 37.8 & 27.3 & 30.2 & 10.8 & 27.3 & 26.5 & 36.9 & 25.4 & 16.9 & 49.8 & 33.3 \\
VideoChat-Online-4B~\citep{huang2025videochatonline} & 28 & 26.2 & 24.0 & 35.9 & 25.1 & 28.3 & 17.2 & 21.0 & 20.6 & 32.0 & 22.7 & 20.9 & 38.6 & 29.8 \\
Flash-VStream-7B~\citep{zhang2025flash} & 26 & 27.1 & 23.8 & 32.7 & 26.3 & 27.6 & 20.2 & 22.3 & 29.1 & 32.7 & 26.1 & 23.3 & 33.8 & 28.5 \\
Mage-VL~\citep{yang2026mage} & 9 & 35.7 & 34.5 & 38.5 & 28.8 & 33.9 & 29.6 & 40.1 & \second{}43.0 & 36.0 & 37.2 & 32.0 & 38.6 & 35.3 \\
RoboBrain2.5-8B~\citep{tan2026robobrain25} & 20 & 30.5 & 32.6 & 34.6 & 26.0 & 31.1 & 28.6 & 26.5 & 29.6 & 36.5 & 30.3 & 20.9 & 39.6 & 30.3 \\
LLaVA-OneVision-2-8B~\citep{an2026llava} & 11 & 34.4 & 30.5 & 37.2 & 25.4 & 31.0 & 34.2 & 35.8 & 37.7 & 39.7 & 36.9 & 31.4 & 38.2 & 34.8 \\
InternVideo3-8B~\citep{yan2026internvideo3} & 18 & 30.9 & 30.3 & 37.2 & 27.9 & 31.8 & 32.5 & 28.6 & 23.3 & 38.8 & 30.8 & 24.4 & 35.3 & 29.8 \\
\addlinespace[2pt]
\midrule
\addlinespace[1pt]
\multicolumn{14}{l}{\textit{Spatial Understanding Models}} & \\
\addlinespace[1pt]
SenseNova-SI-1.5-8B~\citep{sensenova-si} & 23 & 28.8 & 32.8 & 30.8 & 25.2 & 29.6 & 23.9 & 24.1 & 30.0 & 36.0 & 28.5 & 25.0 & 30.9 & 28.0 \\
VST-7B-RL~\citep{yang2025visualspatialtuning} & 25 & 28.4 & 26.8 & 38.5 & 23.3 & 29.5 & 19.5 & 26.3 & 29.6 & 39.4 & 28.7 & 19.8 & 32.9 & 26.3 \\
Cambrian-P-7B~\citep{yang2026cambrianp} & 27 & 26.5 & 28.5 & 32.7 & 19.6 & 26.9 & 9.1 & 28.1 & 30.9 & 26.5 & 23.7 & 24.4 & 38.6 & 31.5 \\
Cambrian-S-7B~\citep{yang2025cambrians} & 31 & 23.8 & 25.1 & 35.9 & 16.2 & 25.7 & 5.4 & 18.8 & 17.5 & 28.3 & 17.5 & 26.2 & 40.6 & 33.4 \\
Spatial-TTT-3B~\citep{liu2026spatialttt} & 32 & 22.0 & 25.5 & 32.1 & 0.0 & 19.2 & 13.1 & 14.1 & 32.7 & 30.4 & 22.6 & 19.8 & 30.0 & 24.9 \\
\addlinespace[2pt]
\midrule
\addlinespace[1pt]
\multicolumn{14}{l}{\textit{Ours}} & \\
\addlinespace[1pt]
Qwen3-VL-8B~\citep{bai2025qwen3vl} & -- &35.5 & 43.1 & 36.5 & 23.6 & 34.4 & 38.4 & 42.7 & 26.9 & 41.5 & 37.4 & 25.6 & 41.5 & 33.6 \\
Qwen3-VL-8B-SFT (Blind) & -- & 29.6 & 26.4 & 32.7 & 25.0 & 28.0 & 23.2 & 23.3 & 32.1 & 34.4 & 28.3 & 34.3 & 35.1 & 34.7 \\
Qwen3-VL-8B-SFT (Ours)$^\dagger$ & -- & 60.4 & 85.6 & 86.5 & 63.4 & 78.5 & 45.3 & 45.6 & 52.5 & 58.5 & 50.5 & 61.6 & 44.4 & 53.0 \\
\bottomrule
\end{tabular}
\label{tab:main}
\end{table}

\paragraph{\textbf{Blind, chance, and frequency baselines.}}
\emph{Blind} evaluates Seed 2.0 Pro with task instructions, questions,
and answer options without any visual input.
\emph{Chance} averages $1/K$ over multiple-choice questions with $K$
options.
\emph{Frequency} predicts the most frequent ground-truth option letter
or numeric value within each question subtype, with scores weighted
by subtype question counts. It is computed from the evaluation answer
distribution to quantify answer imbalance.
All model responses follow the task-specific scoring rules
(Section~\ref{sec:protocol}).

\subsection{Main Results}
The task-level results in Tab.~\ref{tab:main} reveal uneven performance
across the benchmark. Among models evaluated without task-specific fine-tuning, there is a 9.2\% gap between
GPT-5.6 Sol (50.5\%) and Qwen3-VL-32B (41.3\%), the best performing
proprietary and open-source general-purpose models, respectively.
However, rankings varies across tasks.
Mage-VL achieves the highest accuracy on unique-object counting
(T6) among these baselines, at 43.0\%. Its overall accuracy is
35.7\%, the best among video-understanding models but falls behind the general-purpose MLLMs. All evaluated spatial-understanding models perform poorly, with overall accuracies ranging from 22.0\% to 28.8\%.

\paragraph{\textbf{Human Evaluation.}}
We evaluate three independent human annotators on \ourbenchmini{}, with
unrestricted video viewing and replay. Their mean accuracy reaches 90.1\%,
compared with 52.2\% for GPT-5.6 Sol, a difference of 37.9\%.
Human performance remains high across all nine tasks, supporting the benchmark's
solvability.

\begin{table}[t]
\centering
\small
\setlength{\belowcaptionskip}{0pt}
\begin{minipage}[t]{0.485\linewidth}
\centering
\caption{Agentic video understanding frameworks on \ourbenchmini{}(\%). Higher-capability and lightweight backbones are replaced with GPT-5.6 Sol and GPT-5.6 Luna, respectively, with all other configurations unchanged. VA denotes VideoAgent. Assoc., Persist., and Reason.\ average the task accuracies of T1--T3, T4--T7, and T8--T9.}
\label{tab:agentic_mini50}
\end{minipage}\hfill
\begin{minipage}[t]{0.485\linewidth}
\centering
\caption{Frame selection strategies ablation on \ourbenchmini{} (\%)
using GPT-5.6 Sol. 1FPS denotes sampling the video at 1 frame per second and cov.-aware denotes coverage aware sampling to keep max information. Avg is the mean across nine tasks. Assoc., Persist., and Reason.\ average the task accuracies of T1--T3, T4--T7, and T8--T9.}
\label{tab:mini_frame_selection}
\end{minipage}\par
\vspace{6pt}
\nointerlineskip
\begin{minipage}[t]{0.485\linewidth}
\centering
\setlength{\tabcolsep}{2pt}\renewcommand{\arraystretch}{1.08}
\begin{tabular*}{0.95\linewidth}{@{\extracolsep{\fill}}lrrrrr@{}}
\toprule
Task & Chance & Direct & AVP & DVD & VA \\
\midrule
T1 & 25.0 & \textbf{56.0} & 52.0 & 28.0 & 30.0 \\
T2 & 33.3 & \textbf{60.0} & 54.0 & 54.0 & 36.0 \\
T3 & -- & \textbf{48.0} & 22.0 & 30.0 & 32.0 \\
Assoc. & -- & \textbf{54.7} & 42.7 & 37.3 & 32.7 \\
T4 & -- & 44.0 & 32.0 & \textbf{48.0} & 16.0 \\
T5 & -- & \textbf{54.0} & 40.0 & 52.0 & 34.0 \\
T6 & -- & \textbf{40.0} & 32.0 & 18.0 & 18.0 \\
T7 & 25.0 & 56.0 & \textbf{66.0} & 56.0 & 36.0 \\
Persist. & -- & \textbf{48.5} & 42.5 & 43.5 & 26.0 \\
T8 & 25.0 & \textbf{50.0} & 40.0 & 44.0 & 24.0 \\
T9 & 25.0 & 62.0 & 58.0 & \textbf{66.0} & 44.0 \\
Reason. & -- & \textbf{56.0} & 49.0 & 55.0 & 34.0 \\
\midrule
\textbf{Avg} & -- & \textbf{52.2} & 44.0 & 44.0 & 30.0 \\
\bottomrule
\end{tabular*}
\end{minipage}\hfill
\begin{minipage}[t]{0.485\linewidth}
\centering
\setlength{\tabcolsep}{3pt}\renewcommand{\arraystretch}{1.08}
\begin{tabular*}{0.95\linewidth}{@{\extracolsep{\fill}}lrrrr@{}}
\toprule
Task & 1FPS & Uniform & Weighted & Cov.-aware \\
\midrule
T1 & 56.0 & 54.0 & 48.0 & \textbf{62.0} \\
T2 & 60.0 & \textbf{66.0} & 56.0 & \textbf{66.0} \\
T3 & \textbf{48.0} & 26.0 & 44.0 & 44.0 \\
Assoc. & 54.7 & 48.7 & 49.3 & \textbf{57.3} \\
T4 & 44.0 & 50.0 & 36.0 & \textbf{58.0} \\
T5 & 54.0 & \textbf{58.0} & 44.0 & 56.0 \\
T6 & 40.0 & 22.0 & 30.0 & \textbf{42.0} \\
T7 & 56.0 & 52.0 & 56.0 & \textbf{60.0} \\
Persist. & 48.5 & 45.5 & 41.5 & \textbf{54.0} \\
T8 & \textbf{50.0} & 32.0 & 44.0 & 42.0 \\
T9 & 62.0 & 68.0 & \textbf{72.0} & 68.0 \\
Reason. & 56.0 & 50.0 & \textbf{58.0} & 55.0 \\
\midrule
\textbf{Avg}  & 52.2 & 47.6 & 47.8 & \textbf{55.3} \\
\bottomrule
\end{tabular*}
\end{minipage}
\end{table}

\paragraph{\textbf{Agentic Video Understanding.}}
We re-implement AVP~\citep{wang2026avp}, DVD~\citep{zhang2026dvd},
and VideoAgent~\citep{wang2024video} on \ourbenchmini{}.
To control for differences in backbone capabilities, we replace higher-capability models (e.g., Gemini 2.5 Pro, GPT-4 and o3) with GPT-5.6 Sol and lightweight models (e.g., Mini variants) with GPT-5.6 Luna, while retaining all other framework-specific configurations.
Agents can access the full video and revisit segments through their
respective tools.
Under these configurations, none of the three frameworks outperform over direct inference.
AVP and DVD each achieve 44.0\%, while VideoAgent achieves 30.0\%, compared with 52.2\% for GPT-5.6 Sol (Tab.~\ref{tab:agentic_mini50}).

\paragraph{\textbf{Targeted Fine-Tuning.}}
Fine-tuning Qwen3-VL-8B on our separate 30k-question training set
raises overall accuracy from 35.5\% to 60.4\%, a gain of 24.9 percentage points
(Tab.~\ref{tab:main}).
Accuracy improves on all nine tasks, with the largest gains in association
and smaller, less consistent gains in persistent state and reasoning.
The higher ProcTHOR accuracy may partly reflect similar design and furnishing
styles within the dataset, despite no scene, route, or video overlap between
training and evaluation. Average accuracy also improves on 3D-FRONT and
real-world scenes, which are absent from fine-tuning, by 21.9 and 14.8 percentage
points, respectively (Tab.~\ref{tab:appendix_sft_sources}).
These results demonstrate the utility of targeted supervision, while
the uneven gains motivate methods that more consistently maintain
spatial state and use it for reasoning across revisits.

\subsection{Further Analysis}

\begin{figure}[t]
  \centering
  \includegraphics[width=\linewidth]{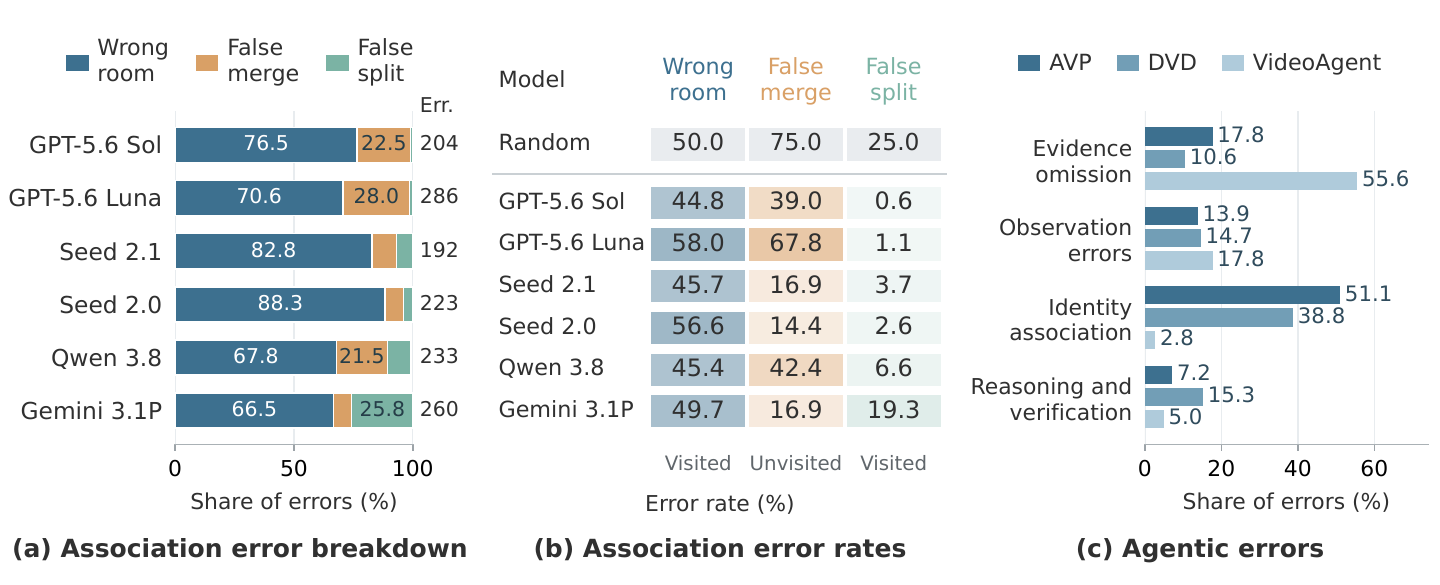}
    \caption{\textbf{Association errors and agentic failure patterns.}
    (a) Error distributions for six proprietary models on 466 T1 task queries,
    normalized by each model's total number of errors (Err.).
    (b) Error rates on visited-room queries ($n=348$; wrong room and
    false split) and unvisited-room queries ($n=118$; false merge).
    \emph{Random} denotes uniform guessing among the options.
    (c) Shares of failures in four major categories on \ourbenchmini{},
    based on manual inspection of agent execution traces.
    Gemini 3.1P denotes Gemini 3.1 Pro.}
  \label{fig:error_analysis}
  \label{fig:t1_errors}
\end{figure}

\paragraph{\textbf{Can models distinguish rooms across revisits?}}
We analyze T1 errors across the six proprietary models, including the strongest zero-shot models. Errors comprise \emph{wrong room} (choosing another visited room), \emph{false merge} (labeling an unentered room as visited), and \emph{false split} (labeling an entered room as unentered). Wrong-room selections dominate, accounting for 66.5--88.3\% of errors (Figure~\ref{fig:error_analysis}(a)). False-split rates on visited-room queries and false-merge rates on unvisited-room queries are below their respective uniform-random baselines of 25.0\% and 75.0\% (Figure~\ref{fig:error_analysis}(b)). However, models still select an incorrect visited room in 44.8--58.0\% of visited-room queries. These results suggest that models can distinguish visited from unvisited rooms to some extent, but often fail to identify which previously visited room the query depicts.

\paragraph{\textbf{Why do agentic frameworks fail?}}
We manually inspect failed trajectories from AVP, DVD, and VideoAgent on \ourbenchmini{}, reviewing their reasoning traces, tool calls, observed frames, and final answers (Fig.~\ref{fig:error_analysis}(c)). We identify four recurring major failure types: missing evidence, inaccurate observations, incorrect room-identity associations, and reasoning or verification errors. Trajectories with multiple problems are classified by the first observable error. AVP and DVD most often make errors in room identity association across revisit, whereas VideoAgent more frequently misses relevant evidence from video sequence. These findings suggest that evidence verification should check not only whether the necessary evidence has been retrieved, but also whether it has been interpreted correctly and linked to the correct room and visit.

\paragraph{\textbf{Does preserving event coverage improve performance?}}
We compare uniform, most relevant event weighted, and event coverage aware sampling on \ourbenchmini{} using GPT-5.6 Sol, with identical prompts and a fixed budget of 128 frames. Both event-based strategies use walkthrough route annotations: \emph{event-weighted sampling} assigns 4$\times$ the sampling weight to relevant room-entry events and \emph{coverage-aware sampling} retains at least one frame from each room-entry, look-around, and hallway transitions interval. Coverage-aware sampling achieves 55.3\%, compared with 47.6\% for uniform and 47.8\% for event-weighted sampling; the separate 1\,fps baseline scores 52.2\% (Table~\ref{tab:mini_frame_selection}). Omitting an intervening bedroom visit, for example, can make two bathroom visits appear continuous. These annotation-guided results suggest that preserving intervening visits and transitions helps distinguish revisits from continuous stays, highlighting the importance of event coverage alongside query-relevant observations.

\section{Conclusion}
We introduced \ourbench{} for evaluating persistent spatial
memory across controlled revisits, together with RS-30K for targeted
model training. Across synthetic and real homes, the benchmark evaluates room association, persistent state tracking, and downstream reasoning under viewpoint variation, identity ambiguity, and state changes.
Our evaluations reveal a substantial human--model gap and no overall
advantage from the evaluated agentic frameworks. The analyses expose
a gap between recognizing a room as familiar and linking it to the
correct spatial memory, while annotation-guided sampling demonstrates
the value of retaining intervening visits and transitions.
Fine-tuning on RS-30K improves Qwen3-VL-8B from 35.5\% to 60.4\%,
with gains extending to scene sources absent from training, but state
tracking and reasoning benefit less consistently than association.
These findings motivate spatial memory mechanisms that preserve room identity across viewpoints, distinguish visually similar rooms, and update memories as room and object states change, while retaining visit histories for downstream reasoning. We hope that \textsc{ReSpaceBench}, RS-30K together with the analysis can provide a useful foundation for systematic evaluation, failure diagnosis, and targeted model development, and ultimately support further research toward more robust and generalizable persistent spatial memory and reasoning across revisits.

\clearpage

\bibliographystyle{plainnat}
\bibliography{main}

\clearpage

\beginappendix

\newif\ifappendixtoc
\addtocontents{toc}{\protect\appendixtoctrue}
\begingroup
\let\appendixtoccontentsline\contentsline
\renewcommand{\contentsline}[4]{\ifappendixtoc\appendixtoccontentsline{#1}{#2}{#3}{#4}\fi}
\renewcommand{\contentsname}{Appendix Contents}
\renewcommand{\cfttoctitlefont}{\sffamily\sectionfont\bfseries\color{seedblue}}
\setcounter{tocdepth}{2}%
\tableofcontents
\endgroup
\clearpage

\section{Implementation Details}
\label{app:implementation}

\subsection{Model Evaluation Settings}
\label{app:video_settings}

We resize video frames to $640\times480$ pixels and initially sample at 1\,fps for default setting for all evaluated models. Due to model-specific context-length limits, we further uniformly
downsample the sampled sequence when necessary, retaining the maximum
number of frames that each model can accommodate alongside the question,
answer options, reference inputs, and reserved response tokens. Sequences
that fit are used in full. All retained frames preserve their chronological order. For multi-home visit questions, the context
budget covers all sequences jointly.

For the models that can't fit our 1\,fps video input, our evaluation use 320 for VST-7B-RL, GLM-4.6V-Flash, and GLM-4.6V (106B);
480 for Gemma-4 E2B, Gemma-4 E4B, Cambrian-S, and Cambrian-P;
150 for InternVL3.5-8B, InternVL3.5-38B, and InternVL3.5-30B-A3B;
and 120 for SenseNova-SI-1.5.

\subsection{Supervised Fine-Tuning}
\label{app:sft_settings}

We fine-tune Qwen3-VL-8B-Instruct on RS-30K, comprising 30,018 questions
from ProcTHOR training scenes. The language backbone is fully updated;
the vision encoder and multimodal projector are frozen. Targets are
the answer option letters or numeric answers.
Table~\ref{tab:appendix_sft_hyperparams} lists the optimization settings.
Both training and testing follow the video-input protocol in
Section~\ref{app:video_settings}, including the resolution, and 1\,fps sampling.

\begin{table}[!htbp]
  \centering
  \caption{Supervised fine-tuning settings for Qwen3-VL-8B.}
  \label{tab:appendix_sft_hyperparams}
  \small
  \setlength{\tabcolsep}{6pt}
  \begin{tabular}{@{}ll@{}}
    \toprule
    \textbf{Hyperparameter} & \textbf{Value} \\
    \midrule
    Training epochs & 1 \\
    Global batch size & 32 \\
    Gradient accumulation & 4 steps \\
    Optimizer & AdamW \\
    Learning rate & $1\times10^{-5}$ \\
    Learning-rate schedule & Cosine decay \\
    Warmup ratio & 0.1 \\
    Adam $(\beta_1,\beta_2)$ & $(0.9,0.95)$ \\
    Weight decay & 0.1 \\
    Gradient clipping (max. norm) & 1.0 \\
    Numerical precision & bfloat16 \\
    Random seed & 42 \\
    \bottomrule
  \end{tabular}
\end{table}

\subsection{Results by Scene Source}
\label{app:sft_sources}

Table~\ref{tab:appendix_sft_sources} compares the base and fine-tuned
Qwen3-VL-8B models on the same 1\,fps evaluation questions, split into
ProcTHOR, 3D-FRONT, and real-world scenes. Each subset's score averages
its task accuracies with equal weight.
Fine-tuning uses only ProcTHOR training scenes and improves the average on all three sources, including the real scenes.

\begin{table}[!htbp]
  \centering
  \caption{Per-task average accuracy (\%) of base and SFT Qwen3-VL-8B at 1\,fps, grouped by scene source.$\Delta$ is the SFT gain in percentage points.}
  \label{tab:appendix_sft_sources}
  \small
  \setlength{\tabcolsep}{6pt}
  \begin{tabular}{@{}lrrrr@{}}
    \toprule
    \textbf{Scene source} & \textbf{Questions} & \textbf{Base Avg.} & \textbf{SFT Avg.} & $\boldsymbol{\Delta}$ \\
    \midrule
    ProcTHOR & 1,471 & 34.7 & 74.5 & $+39.8$ \\
    3D-FRONT & 1,555 & 33.2 & 55.1 & $+21.9$ \\
    Real-world & 218 & 40.4 & 55.2 & $+14.8$ \\
    \midrule
    All & 3,244 & 35.5 & 60.4 & $+24.9$ \\
    \bottomrule
  \end{tabular}
\end{table}

\clearpage
\section{Data Construction and Question Definitions}
\label{app:construction_details}

\paragraph{\textbf{Room entries and persistent identity.}}
Room entries are recovered from the rendered camera trajectory and its room
labels. A visit requires entering a room; viewing it through a doorway does
not count. Re-entry into the same physical room creates a new visit while
preserving its identity. The initial countable room is included as the first
visit, and hallways, corridors, stairs, and other transit spaces are excluded.
Room inventories and first-encounter ordinals count each physical room once,
whereas revisit counts retain separate entry events.

\paragraph{\textbf{Ground-truth verification.}}
Answers are recomputed from the verified room-entry sequence and object-state
records. For entry-dependent questions, we check the annotated entry order
against the recorded look-arounds. Questions with inconsistent annotations or
ambiguous room references are excluded. Object counts use persistent instance
identities to avoid counting the same object again on a revisit.

\paragraph{\textbf{Task-specific evidence validation.}}
To ensure disambiguity in synthetic scene annotations, room labels are checked against the furnishings and fixtures
present after scene preparation. T1 (query-to-memory association) verifies
whether the query room was entered, and T9 (temporal successor) uses the
realized order of countable-room entries. Object related questions additionally
applies geometric visibility and image-quality checks to the
queried objects over multiple frames, with further framing and occlusion
checks for reference images. Only visible objects in the video are counted towards the ground-truth.

\clearpage
\subsection{Task Subtypes and Templates}
\label{app:question_subtypes}
Table~\ref{tab:question_subtypes} summarizes the task definitions and gives
representative question wording from the evaluation set. Task numbers and
names follow the taxonomy in Section~\ref{sec:tasks}.

\begin{table}[!htbp]
\centering\small
\setlength{\tabcolsep}{4pt}
\renewcommand{\arraystretch}{1.12}
\begin{tabular}{@{}>{\raggedright\arraybackslash}p{.24\linewidth}>{\raggedright\arraybackslash}p{.25\linewidth}>{\raggedright\arraybackslash}p{.44\linewidth}@{}}
\toprule
Task & Subtypes / target & Representative template \\
\midrule
T1\newline Query-to-memory association & Query clip matched to a previously entered room, or never entered (MCQ).
& The video above shows one or more walks through homes. Did you walk into this room in the short clip during your walk(s)? If so, which one was it? \\
\addlinespace[5pt]
T2\newline Same-instance identification & Current room matched to an earlier room of the same type (MCQ).
& Which of the rooms you entered earlier is the one you are in now? \\
\addlinespace[5pt]
T3\newline Revisit counting & Prior entries into the current physical room (numeric).
& In the room you're currently in, how many times have you been to THIS EXACT room before? \\
\addlinespace[5pt]
T4\newline Room inventory & Distinct-room count by type; first-encounter ordinal (numeric).
& You walked through a home. How many different bathrooms did you walk into? / Number the distinct rooms in the order you first walked into them. Which number is the kitchen? \\
\addlinespace[5pt]
T5\newline Object--place binding & First room containing an object category, in distinct-room order (numeric).
& Number the distinct rooms in the order you first walked into them. What is the number of the first room you entered with a refrigerator in it? \\
\addlinespace[5pt]
T6\newline Unique-object counting & Unique object instances across entered rooms and revisits (numeric).
& How many different sofas did you see in total for the rooms you entered? \\
\addlinespace[5pt]
T7\newline Cross-session state & First-to-second visit change; first/last locations; last observed room before disappearance; four-visit location history (MCQ).
& The picture shows a chest of drawers you saw on your first visit. Comparing your first visit with your second visit, what happened to it? / Where was a potted plant in the living room on each of your four visits? \\
\addlinespace[5pt]
T8\newline Spatial topology & Room pair directly connected by a doorway (MCQ).
& Which two of these rooms are directly connected to each other by a doorway (with no other room or passages in between)? \\
\addlinespace[5pt]
T9\newline Temporal successor & Next countable room type after a specified entry (MCQ).
& Right after you left the kitchen for the first time, which kind of room did you walk into next? (Hallways and other passages you just walk through never count.) \\
\bottomrule
\end{tabular}
\caption{Tasks and representative evaluation templates.
The real-video cross-session extension also asks which objects are newly
present on the second visit.}
\label{tab:question_subtypes}
\end{table}

\subsection{Camera Motion}
\label{app:construction_additional}

\begin{table}[!htbp]
\centering
\caption{Camera motion statistics of the rendered walkthroughs.}
\label{tab:motion}
\small
\setlength{\tabcolsep}{8pt}
\begin{tabular}{@{}lcc@{}}
\toprule
Metric & Mean & Range or nominal setting \\
\midrule
Walking speed during movement & 0.662\,m/s & 0.6--0.8\,m/s \\
Average traversal speed, including pauses & 0.372\,m/s & 0.227--0.503\,m/s \\
Camera displacement between frames at 1\,fps & 0.367\,m & 0--0.833\,m \\
\bottomrule
\end{tabular}
\end{table}

\paragraph{\textbf{Camera motion and visit coverage at 1\,fps.}}
Table~\ref{tab:motion} summarizes the camera motion in the rendered
walkthroughs. At 1\,fps, consecutive sampled frames are 0.367\,m apart on
average. The exact instant of entering or leaving a room may fall between
two sampled frames, but temporally adjacent observations remain available.
Moreover, each counted visit includes an in-room look-around lasting 6--9
seconds, which provides approximately 6--9 sampled views under the full
1\,fps setting.

\subsection{Shortcut Controls During Construction}
\label{app:shortcut_controls}

The construction pipeline includes several checks to reduce non-visual
shortcuts~\citep{vsibench_debiased}. At the question level, the question text does not
state information that must be recalled from the video. We balance
the option letters (A/B/C/D) as well as the answer content, such as
the room type (bathroom/bedroom/living room) or the ordinal option (first/second/not entered). Questions with the same
wording and the same options are paired with different visual inputs that
lead to different answers, so that answering depends on the video.
Distractors are rooms of the same type that appear in the same video, so they cannot be
excluded by common sense.

We further check for
priors and textual shortcuts by testing strategies that guess a fixed answer
from the room type or object category, or that answer from option length,
lexicographic order, or ordinal words. After rendering, visibility, and
quality filtering, we repeat these checks on the resulting answer distributions, so
that filtering does not reintroduce biases.

\clearpage

\section{Additional Analyses}
\label{app:additional}

\subsection{Visual-Input Diagnostic}
\label{app:visual_input}

\begin{table}[!htbp]
\centering
\caption{Effect of the visual input on the fine-tuned Qwen3-VL-8B, on all
3,244 \ourbench{} questions.}
\label{tab:visual_input}
\small
\setlength{\tabcolsep}{8pt}
\begin{tabular}{@{}lc@{}}
\toprule
Visual-input condition & Nine-task average (\%) \\
\midrule
Correct visual input & 60.4 \\
No visual input (blind) & 29.6 \\
Randomly mismatched visual input & 20.8 \\
\bottomrule
\end{tabular}
\end{table}

To further test whether the fine-tuned model relies on textual shortcuts or on visual evidence, we evaluate
the same checkpoint on all 3,244 questions under three conditions with
matched evaluation settings: the correct visual input, no visual input
(\emph{blind}), and a randomly mismatched visual input. In the mismatched
condition, the visual input of each question is replaced with that of a
randomly sampled question from a different scene whose ground-truth answer
differs. As shown in Tab.~\ref{tab:visual_input}, removing the visual input reduces
accuracy by 30.8 points, and mismatched inputs reduce it by a further 8.8
points. The fine-tuned model therefore relies substantially on visual
evidence rather than on textual shortcuts.

\subsection{Does Identity Association Limit Cross-Session State Tracking?}
\label{app:t7_oracle}

\begin{table}[!htbp]
\centering
\caption{Identity-oracle analysis on T7 (cross-session state) with GPT-5.6
Sol on all 559 T7 questions. Gains are in percentage points over the
original input.}
\label{tab:t7_oracle}
\small
\setlength{\tabcolsep}{8pt}
\begin{tabular}{@{}lcc@{}}
\toprule
Visual input & Accuracy (\%) & Gain \\
\midrule
Original input & 47.0 & -- \\
+ Room type and persistent room ID & 51.9 & $+4.9$ \\
+ Object masks and consistent instance IDs & 54.0 & $+7.0$ \\
\bottomrule
\end{tabular}
\end{table}

We run an identity-oracle analysis on cross-session state (T7). The task requires identifying the room and the object before comparing their
states. To test whether identity association limits performance, we evaluate GPT-5.6 Sol on all 559 T7 questions while
providing ground-truth identity (Tab.~\ref{tab:t7_oracle}). The
\emph{room oracle} annotates each frame with its room type and a unique room
ID. The \emph{room + object oracle} additionally overlays object masks with
unique instance IDs onto the frames. These annotations provide identity correspondence
without revealing object states, state changes, or ground-truth answers.
Providing room IDs improves accuracy by 4.9 points, and
adding object-level annotations yields a total gain of 7.0 points. These results suggest that identity correspondence is one source of difficulty in cross-session state-change reasoning, while substantial errors remain even when ground-truth identity cues are provided.

\subsection{Generalization Beyond \ourbench{}}
\label{app:ood}

\begin{table}[!htbp]
\centering
\caption{Out-of-distribution generalization of Qwen3-VL-8B under different
fine-tuning data configurations (\%).}
\label{tab:ood_results}
\small
\setlength{\tabcolsep}{6pt}
\begin{tabular}{@{}lccc@{}}
\toprule
Benchmark & \makecell{Original\\Qwen3-VL-8B} & \makecell{VLM-3R +\\LLaVA-Video} & \makecell{VLM-3R +\\LLaVA-Video + RS-30K} \\
\midrule
VSI-Bench & 58.2 & 64.3 & 66.7 \\
VSI-Bench-Debiased & 53.2 & 55.9 & 57.0 \\
Video-MME (64 frames) & 66.1 & 67.0 & 67.6 \\
\midrule
\ourbench{} & 35.5 & 23.1 & 63.6 \\
\bottomrule
\end{tabular}
\end{table}

Improving \ourbench{} alone does not fully establish generalization.
Fine-tuning on RS-30K alone substantially improves \ourbench{} but reduces
performance on some external benchmarks, a limitation of training
exclusively on task-specific data. We therefore study mixed-data
fine-tuning, following the common practice of combining specialized and
general instruction data in spatial MLLM training, as in
MM-Spatial~\citep{mmspatial} and SpatialRGPT~\citep{spatialrgpt}. With the same
Qwen3-VL-8B~\citep{bai2025qwen3vl} backbone, we fine-tune on partial VLM-3R~\citep{vlm3r} and
LLaVA-Video~\citep{llavavideo} data, with and without RS-30K.

Table~\ref{tab:ood_results} reports the results. Adding RS-30K improves
VSI-Bench~\citep{yang2025thinking} and its debiased
subset~\citep{vsibench_debiased} by 2.4 and 1.1 points, respectively, and raises
\ourbench{} accuracy to 63.6\%, a gain of 28.1 points over the original
model. Video-MME~\citep{videomme} reaches 67.6\%, above both the mixture
without RS-30K and the original model. Combined with broader training data,
RS-30K thus provides spatial-memory supervision without sacrificing
performance on the evaluated external benchmarks.

\subsection{Trajectory Complexity}
\label{app:trajectory}

\begin{figure}[!htbp]
\centering
\includegraphics[width=\linewidth]{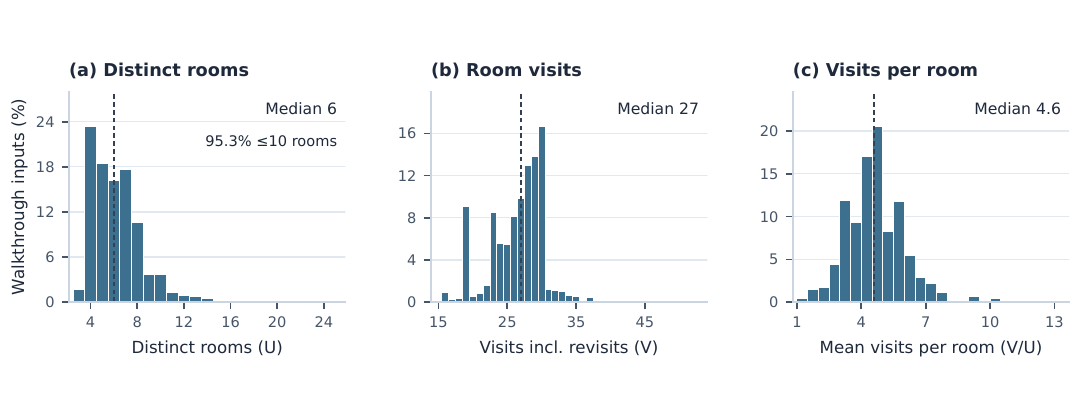}
\caption{Trajectory statistics of \ourbench{}: percentage of walkthrough inputs by (a) distinct rooms $U$, (b) room visits $V$ including revisits, and (c) visits per room $V/U$. Dashed lines mark the medians.}
\label{fig:traj_distributions}
\end{figure}

\begin{figure}[!htbp]
\centering
\includegraphics[width=0.8\linewidth]{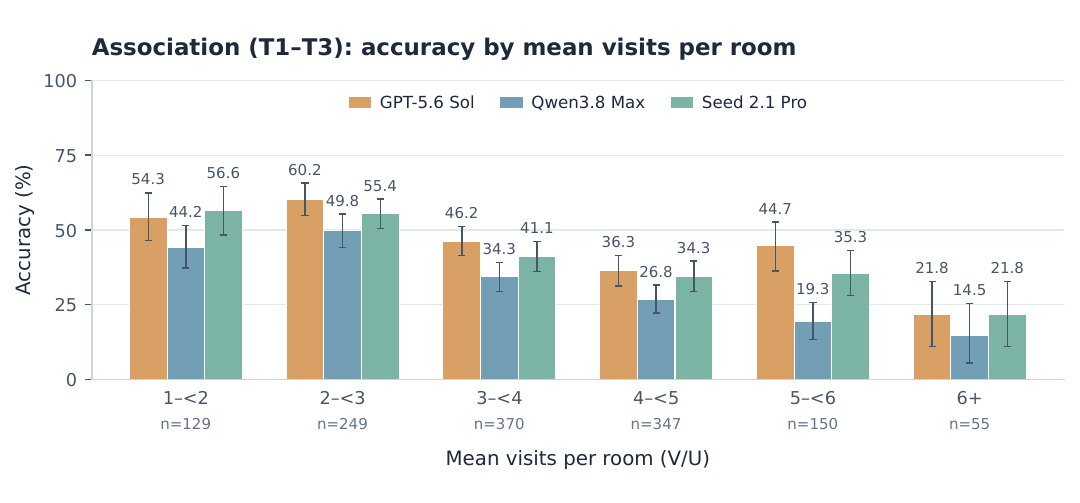}
\caption{Association accuracy by mean visits per room ($V/U$), mean over all T1--T3 questions in each bin; $n$ is the number of questions.}
\label{fig:assoc_visits_per_room}
\end{figure}

Figure~\ref{fig:traj_distributions} shows the statistics of
(a) distinct rooms, (b) room visits, and (c) visits per room over the
walkthrough inputs of \ourbench{}. The median number of distinct rooms per
input is 6, and 95.3\% of inputs involve at most
10 distinct rooms. The median number of room visits, including revisits, is
27. Rooms are therefore entered repeatedly: the
median visits-per-room ratio is 4.6.

We further relate the association-family accuracy of three proprietary models to the mean number of visits per room. Figure~\ref{fig:assoc_visits_per_room} relates association accuracy
(T1--T3) to the visits-per-room ratio for GPT-5.6 Sol, Qwen 3.8 Max, and
Seed 2.1 Pro. All three models are most accurate when rooms are visited
fewer than three times on average (44.2--60.2\%). Accuracy generally
declines as rooms are revisited more often and reaches 14.5--21.8\% when
rooms are visited six or more times on average. This suggests that frequent
revisits make room association harder.

\subsection{Evaluation Under a Common Frame Budget}
\label{app:common_budget}

\begin{table}[!htbp]
\centering
\caption{Per-task accuracy (\%) on \ourbenchmini{} under the default
protocol and under a common budget of 128 uniformly sampled frames.
$^{*}$The 1\,fps input of this model is capped by its context limit
(Appendix~\ref{app:video_settings}).}
\label{tab:common_128}
\small
\setlength{\tabcolsep}{5pt}
\begin{tabular}{@{}lcccccccccc@{}}
\toprule
Model & Avg. & T1 & T2 & T3 & T4 & T5 & T6 & T7 & T8 & T9 \\
\midrule
\multicolumn{11}{@{}l}{\textit{Default protocol (1\,fps)}} \\
GPT-5.6 Sol & 52.22 & 56 & 60 & 48 & 44 & 54 & 40 & 56 & 50 & 62 \\
Seed 2.0 Pro & 51.56 & 64 & 60 & 30 & 52 & 56 & 22 & 48 & 48 & 84 \\
Seed 2.1 Pro & 43.11 & 60 & 56 & 22 & 36 & 46 & 20 & 48 & 38 & 62 \\
Qwen3-VL-8B & 41.11 & 48 & 48 & 42 & 46 & 52 & 24 & 42 & 24 & 44 \\
InternVL3.5-38B$^{*}$ & 39.11 & 46 & 36 & 34 & 30 & 32 & 36 & 48 & 30 & 60 \\
Gemini 3.1 Pro & 36.89 & 34 & 40 & 30 & 30 & 36 & 24 & 54 & 38 & 46 \\
GPT-5.6 Luna & 32.00 & 38 & 44 & 14 & 26 & 38 & 30 & 34 & 26 & 38 \\
GLM-4.6V-Flash-9B$^{*}$ & 29.11 & 48 & 30 & 28 & 20 & 12 & 16 & 38 & 26 & 44 \\
SenseNova-SI-1.5-8B$^{*}$ & 29.11 & 18 & 22 & 38 & 32 & 18 & 32 & 34 & 30 & 38 \\
VST-7B-RL$^{*}$ & 28.67 & 30 & 38 & 30 & 22 & 20 & 22 & 38 & 26 & 32 \\
Cambrian-S-7B$^{*}$ & 25.11 & 28 & 30 & 22 & 8 & 18 & 16 & 28 & 26 & 50 \\
\midrule
\multicolumn{11}{@{}l}{\textit{Common budget (128 uniformly sampled frames)}} \\
GPT-5.6 Sol & 47.56 & 54 & 66 & 26 & 50 & 58 & 22 & 52 & 32 & 68 \\
Seed 2.0 Pro & 44.89 & 52 & 44 & 24 & 50 & 48 & 24 & 42 & 40 & 80 \\
Seed 2.1 Pro & 39.11 & 54 & 32 & 6 & 40 & 50 & 28 & 48 & 36 & 58 \\
Qwen3-VL-8B & 35.56 & 44 & 30 & 30 & 40 & 38 & 24 & 48 & 24 & 42 \\
InternVL3.5-38B & 39.33 & 48 & 40 & 32 & 30 & 30 & 36 & 48 & 30 & 60 \\
Gemini 3.1 Pro & 34.89 & 56 & 46 & 12 & 18 & 36 & 20 & 46 & 32 & 48 \\
GPT-5.6 Luna & 31.33 & 34 & 42 & 18 & 28 & 28 & 14 & 44 & 28 & 46 \\
GLM-4.6V-Flash-9B & 28.44 & 52 & 30 & 24 & 20 & 10 & 12 & 38 & 26 & 44 \\
SenseNova-SI-1.5-8B & 30.22 & 26 & 24 & 40 & 36 & 24 & 22 & 32 & 30 & 38 \\
VST-7B-RL & 29.11 & 34 & 38 & 30 & 22 & 20 & 24 & 36 & 26 & 32 \\
Cambrian-S-7B & 26.67 & 28 & 36 & 28 & 8 & 20 & 16 & 22 & 30 & 52 \\
\bottomrule
\end{tabular}
\end{table}

Under the default protocol, models with shorter context limits receive fewer
frames (Appendix~\ref{app:video_settings}). To separate model capability from
the amount of visual input, we additionally evaluate 11 representative models
on \ourbenchmini{} with the same 128 uniformly sampled frames
(Tab.~\ref{tab:common_128}). SenseNova-SI-1.5-8B uses $420\times420$ to fit all 128 frames within
its visual token budget. The ranking remains largely consistent: GPT-5.6 Sol
and Seed 2.0 Pro retain the top two positions, and the three spatially
specialized models remain near the bottom. Some intermediate ranks change,
but the relatively poor performance of spatially specialized models cannot
be attributed solely to differences in frame count.

\end{document}